\documentclass[11pt]{article}

\usepackage[preprint]{acl}

\usepackage{times}
\usepackage{latexsym}
\usepackage{adjustbox}

\usepackage[T1]{fontenc}

\usepackage[utf8]{inputenc}

\usepackage{microtype}

\usepackage{inconsolata}
\usepackage{hyperref}
\usepackage{url}

\usepackage{array} 
\newcolumntype{C}[1]{>{\centering\arraybackslash}m{#1}}

\definecolor{rowgray}{RGB}{248,248,248}

\newcommand{\good}{\raisebox{0.2ex}{\normalsize\cmark}}
\newcommand{\bad}{\raisebox{0.2ex}{\normalsize\xmark}}
\newcommand{\limi}{\raisebox{0.2ex}{\normalsize\limited}}

\usepackage{graphicx}
\usepackage{xcolor}
\usepackage{pifont}
\usepackage{booktabs}
\usepackage{subcaption}

\newcommand{\cmark}{\textcolor{green!60!black}{\ding{51}}}
\newcommand{\xmark}{\textcolor{red}{\ding{55}}}
\newcommand{\limited}{\textcolor{yellow!50!orange}{\raisebox{-0.4ex}{\LARGE$\sim\mkern-14.5mu\sim$}}}

\usepackage{algorithm}
\usepackage{algorithmic}
\usepackage{amsmath}
\usepackage{amsthm}
\usepackage[utf8]{inputenc}
\usepackage[T1]{fontenc}
\usepackage{url}
\usepackage{pifont}
\usepackage{booktabs}
\usepackage{amsfonts}
\usepackage{siunitx}
\usepackage{tabularx}
\usepackage{nicefrac}
\usepackage{microtype}
\usepackage{xcolor}
\usepackage{graphicx}
\usepackage{wrapfig}      
\usepackage{caption}      
\usepackage{enumitem}
\setlist[itemize]{left=2pt, itemsep=-2pt, topsep=0pt}

\definecolor{rowgray}{RGB}{248,248,248}
 
\def\redc{\cellcolor[HTML]{FF999A}}
\def\orangec{\cellcolor[HTML]{FFCC99}}
\def\yellowc{\cellcolor[HTML]{FFF8AD}}
\usepackage[table]{xcolor}
\definecolor{lightyellow}{RGB}{255, 255, 224} 

\newcommand{\ours}{PRL}

\usepackage{tikz}
\usetikzlibrary{shapes.geometric, arrows.meta, positioning}
\usepackage[most]{tcolorbox}
\usepackage{multicol}
\usepackage{arydshln}
\usepackage{bibentry}

\tikzset{
  block/.style={
    rectangle,
    draw,
    rounded corners=3pt,
    fill=blue!10,
    align=center,
    minimum width=2.5cm,
    minimum height=1cm,
    font=\small
  },
  arrow/.style={
    -{Latex},
    line width=1pt
  },
  dot/.style={
    font=\Large,
    inner sep=0pt
  }
}

\usepackage{cuted}
\usepackage[textsize=tiny]{todonotes}

\title{PRL: \underline{P}rompts from \underline{R}einforcement \underline{L}earning}

\author{
\begin{tabular}{@{}c@{\hspace{2.5em}}c@{\hspace{2.5em}}c@{}}
Pawe{\l} Batorski$^{*}$ & Adrian Kosmala$^{*}$ & Paul Swoboda \\
\multicolumn{3}{c}{{\normalfont Heinrich Heine Universit\"at D\"usseldorf}} \\
\multicolumn{3}{c}{{\normalfont\texttt{\{pawel.batorski,adrian.kosmala,paul.swoboda\}@hhu.de}}} \\
\multicolumn{3}{c}{{\normalfont $^{*}$Equal contribution.}}
\end{tabular}
}

\begin{document}
\maketitle
\begin{strip}
\vspace{-2cm}

\centering
\small
\setlength{\tabcolsep}{6pt}
\renewcommand{\arraystretch}{1.2}
\noindent
\begin{minipage}[t]{0.42\textwidth}\vspace{0pt}
  \centering
  \includegraphics[width=\linewidth]{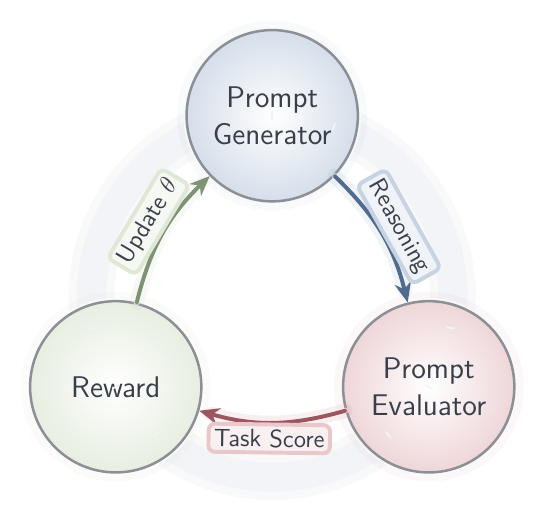}
\end{minipage}\hfill%
\begin{minipage}[t]{0.56\textwidth}\vspace{0pt}
  \centering
  \vspace{1.2cm}
  \begin{tabular}{@{}l*{3}{C{2.2em}}@{}}
    \toprule
    \textbf{Method} & \textbf{Gen.} & \textbf{Ref.} & \textbf{Few-shot} \\
    \midrule
    Manual Instr. & \bad & \bad & \bad \\
    APE           & \good & \bad & \bad \\
    EvoPrompt     & \good & \good & \bad \\
    APO           & \good & \good & \limi \\
    PromptAgent           & \good & \good & \limi \\
    \ours{}       & \good & \good & \good \\
    \bottomrule
  \end{tabular}

  \vspace{3pt}
  {\scriptsize \good\ supported \quad \bad\ not supported \quad \limi\ limited}
\end{minipage}

\captionof{figure}{
    \underline{Left}: Our RL-based prompt optimization cycle (overview).
    \underline{Right}: Comparison of prompt-engineering methods. \ours{} automates both prompt generation and refinement and, synthesizes novel task-specific few-shot examples. The yellow tilde (\limited) for APO and PromptAgent indicates limited few-shot support, its examples are drawn from training data, which restricts performance, whereas \ours{} creates new instances not seen during training.
}
\label{fig:cycle_and_comparison}
\end{strip}

\begin{abstract}
Effective prompt engineering remains a central challenge in fully harnessing the capabilities of LLMs. While well-designed prompts can dramatically enhance performance, crafting them typically demands expert intuition and a nuanced understanding of the task. Moreover, the most impactful prompts often hinge on subtle semantic cues, ones that may elude human perception but are crucial for guiding LLM behavior. In this paper, we introduce \ours{} (Prompts from Reinforcement Learning), a novel RL-based approach for automatic prompt generation. 
Unlike most prior methods, \ours{} can also generate novel few-shot demonstrations rather than selecting only from examples seen during training. Empirically, our approach achieves state-of-the-art performance across a diverse set of benchmarks, including text classification, text simplification, and mathematical reasoning, while remaining competitive on text summarization. Our code is available at \href{https://github.com/Batorskq/prl}{https://github.com/Batorskq/prl}.
 

\end{abstract}

\section{Introduction}
Prompt engineering is a lightweight way to steer LLM behavior without fine-tuning \citep{sahoo2024systematic, chen2023unleashing}. 
Yet, performance can be highly sensitive to seemingly minor, semantics-preserving prompt changes and formatting details \citep{razavi2025benchmarking, sclar2024quantifyinglanguagemodelssensitivity, chatterjee2024posixpromptsensitivityindex, zhuo2024prosaassessingunderstandingprompt, errica2024did, cao2024worstpromptperformancelarge}. 
This brittleness extends to realistic perturbations (e.g., typos/paraphrases) and even instruction-following robustness under injected instructions \citep{zhu2024promptrobustevaluatingrobustnesslarge, li2023evaluatinginstructionfollowingrobustnesslarge}. 
Classic in-context learning results further show strong dependence on demonstration choice and ordering \citep{zhao2021calibrateuseimprovingfewshot, lu2022fantasticallyorderedpromptsthem, min2022rethinkingroledemonstrationsmakes, reynolds2021prompt}, and recent reasoning-focused models also explicitly note prompt sensitivity \citep{guo2025deepseek}.
\noindent
Few-shot prompting, in which a prompt includes a small set of input-output examples, is widely used approach to guide LLMs. While often beneficial, \citep{reynolds2021prompt} demonstrate that zero-shot prompting can sometimes outperform few-shot approaches, suggesting that the usefulness of examples may depend on task familiarity or pretraining exposure. These findings collectively highlight the challenge of designing effective prompts and motivate the need for automated, task-specific prompt optimization. 
\noindent
Recent work has explored automatic prompt engineering and optimization \citep{guo12connecting, zhang2022opt,yang2023large, pryzant2023automatic, wang2023promptagent, shi2025lossgaingatedrefinement}. 
However, most methods, with partial exceptions \citep{pryzant2023automatic, wang2023promptagent}, do not incorporate tailored few-shot demonstrations, and when they do, examples are typically restricted to the training set. 
We propose \ours{}, an RL-based prompt optimization method that learns not only how to refine instructions, but also whether to use few-shot examples and how to synthesize them to maximize task performance. 
Additional discussion and future directions are provided in Appendix~\ref{sec:appendix_discussions}.
\noindent
Interestingly, the incorporation of few-shot examples is spontaneously emerging during the prompt generation training and is not explicitly encouraged.
Additionally, \ours{} incorporates a reasoning phase prior to prompt generation, where the model first produces a rationale to guide its final output. To summarize, our contributions are as follows:
\noindent
\begin{description}[leftmargin=!, labelwidth=10pt]

\item[Conceptual:] We introduce \ours{}, automated prompt-optimization method that can both generate and select unseen few-shot examples.
  \item[Experimental:] We demonstrate strong performance across classification, summarization, simplification, domain-knowledge, and mathematical reasoning benchmarks, and we provide comprehensive ablations.
  \item[Finding:] We find that few-shot prompting behavior emerges naturally during RL training, despite not being explicitly incentivized.
\end{description}

\section{Related Work}

\textbf{Prompt Engineering} steers LLMs at inference time without parameter updates~\citep{sahoo2024systematic, chen2023unleashing}. 
Reasoning-focused prompting includes Chain-of-Thought (CoT)~\citep{wei2022chain}, Tree-of-Thought (ToT)~\citep{yao2023tree}, and structured variants such as Program-of-Thoughts~\citep{chen2022program} and Graph-of-Thoughts~\citep{besta2024graph}. 
Complementary \emph{soft prompting} methods optimize continuous prompts, e.g., Prefix-Tuning~\citep{li2021prefix} and Prompt Tuning~\citep{lester2021power}.

Few-shot prompting (in-context learning)~\citep{brown2020language} conditions models on demonstrations and is effective across tasks~\citep{xu2023llms, greenblatt2024getting, sivarajkumar2024empirical}, but its gains are sensitive to demonstration choice and ordering~\citep{zhao2021calibrateuseimprovingfewshot, lu2022fantasticallyorderedpromptsthem, min2022rethinkingroledemonstrationsmakes} and can even hurt performance~\citep{reynolds2021prompt}. 
Our method learns when to use few-shot examples and how to construct them based on task feedback.

\paragraph{Automated Prompt Engineering}
Automated prompt engineering replaces manual prompt design with algorithmic generation and refinement. RLPrompt uses RL but is limited to very short, often non-interpretable prompts \citep{deng2022rlprompt}, while APE generates natural-language candidates and selects by validation performance \citep{zhou2022large}. Many methods refine prompts via critique/feedback (APO \citep{pryzant2023automatic}, PromptAgent \citep{wang2023promptagent}, GRACE \citep{shi2025lossgaingatedrefinement}, CriSPO \citep{he2025crispo}, ZERA \citep{yi2025zera}) but, when using few-shot demonstrations, typically draw them from training data. EvoPrompt searches via evolutionary operators \citep{guo12connecting}, and GPS targets per-sample prompting without task-specific fine-tuning \citep{batorski2025gpsgeneralpersampleprompter}. In contrast, \ours{} learns whether to use few-shot prompting and can synthesize novel task-specific examples, enabling broader prompt-space exploration.

\paragraph{Multi-agent prompt optimization.}
Some methods optimize prompts for multi-stage LM programs or agentic pipelines rather than a single LLM call, including DSPy~\citep{dspy} (teleprompting), MiPRO~\citep{mipro} and GePA~\citep{gepa} (optimizing instructions and demonstrations across modules), BetterTogether~\citep{bettertogether} (prompt optimization with parameter updates in modular pipelines), and SGLang~\citep{sglang} (efficient execution of structured LM programs). We do not compare to these methods because they target end-to-end multi-module behavior, whereas \ours{} outputs a single reusable prompt for one LLM call.

\begin{figure*}[t]
  \centering
  \includegraphics[width=\linewidth]{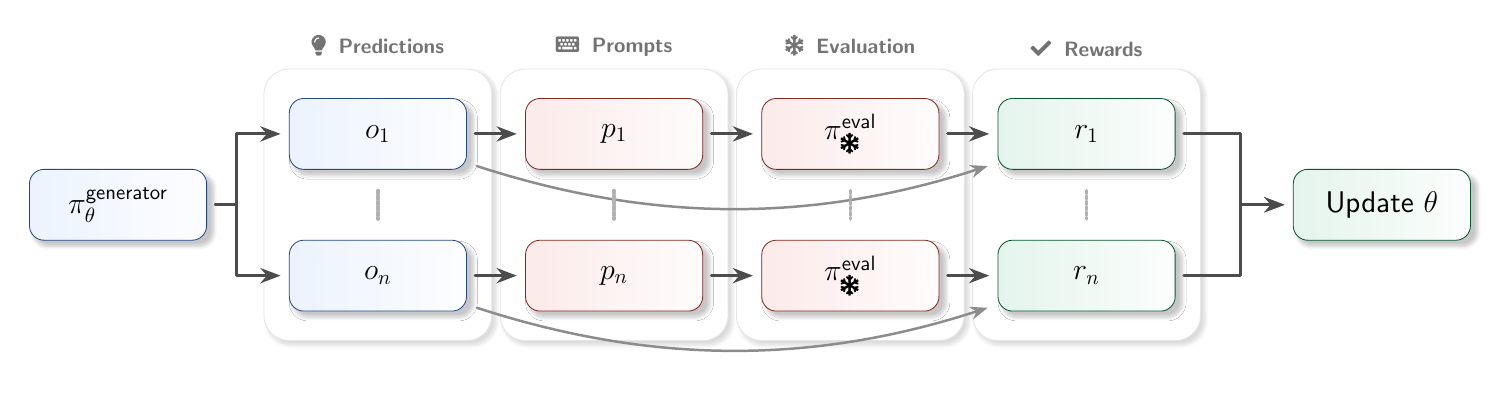}
  \caption{Training scheme of \ours{}. First, the Prompt Generator $\pi_{\theta}^{\text{generator}}$ generates a set of outputs $o_1, \ldots, o_n$ (reasoning + generated prompt) from which the corresponding prompts $p_1, \ldots, p_n$ are extracted. Each prompt is then evaluated by the Evaluation Model $\pi^{\text{eval}}$ (a language model with frozen parameters), which produces corresponding answers. These answers, along with the outputs from the Prompt Generator, are used to compute rewards $r_1, \ldots, r_{n}$. Finally, the rewards are used to update the parameters of the Prompt Generator through RL.}
  \label{fig:training}
\end{figure*}

\section{Method}
Our method comprises the following components:
\begin{itemize}
\item \textbf{Prompt Generator:} A trainable language model that generates prompts with help of a reasoning process, see the prompt formats in Appendix~\ref{sec:prompts_format}.
\item \textbf{Evaluation Model:} A frozen LLM that produces an answer based on the generated prompt. 
\item \textbf{Prompt Selection:} We learn the prompt generator through RL with a reward incorporating both formatting and task performance. We choose the best overall prompt by regularly querying the prompt generator model for prompt candidates and evaluate those.
\end{itemize}

\noindent
We now provide a detailed description of each component of our method.

\paragraph{Reward Function.}
Our reward combines a formatting reward for the Prompt Generator and a task reward from the Evaluation Model:
\[
R \;=\; R_{\text{gen}} \;+\; R_{\text{eval}} .
\]

\paragraph{Prompt Generator reward.}
We define
\[
R_{\text{gen}} \;=\; R_{\text{token}} \;+\; R_{\text{structure}} .
\]

\noindent
\underline{Token usage.} Let $\mathcal{T}=\{\texttt{<think>},\,\texttt{</think>},\,\texttt{<answer>},\,\texttt{</answer>}\}$ be the set of required delimiter tokens, and let $c(\tau)\in\{0,1\}$ indicate whether token $\tau$ appears \emph{exactly once}. Then
\[
R_{\text{token}}
\;=\;
\frac{r_{\text{token}}}{|\mathcal{T}|}
\sum_{\tau \in \mathcal{T}} c(\tau),
\]
so the full reward $r_{\text{token}}$ is obtained only if all tokens are used correctly.

\noindent
\underline{Structure.} We set $R_{\text{structure}} = r_{\text{structure}}$ if the output matches
\texttt{<think>} reasoning \texttt{</think>} \texttt{<answer>} answer \texttt{</answer>}
exactly, and $0$ otherwise.

\paragraph{Evaluation Model reward.}
To measure the quality of a generated prompt, we reward the Evaluation Model based on (i) output validity and (ii) task success:
\[
R_{\text{eval}} \;=\; R_{\text{format}} \;+\; R_{\text{alignment}} .
\]
\noindent
\underline{Format.} We set $R_{\text{format}} = r_{\text{format}}$ if the output satisfies the task-specific format constraints
(e.g., it is a valid label from the predefined class set), and $0$ otherwise. When no format constraint is required, we use
$r_{\text{format}}=0$.
\noindent \newline
\underline{Alignment.} We set $R_{\text{alignment}} = r_{\text{alignment}}$ if the output is correct or achieves the task objective,
where correctness is measured by the task metric (e.g., accuracy, depending on the task); otherwise $0$.

\noindent
Overall, the total reward is
\[
R \;=\; R_{\text{token}} + R_{\text{structure}} + R_{\text{format}} + R_{\text{alignment}} ,
\]
with task-specific definitions of $R_{\text{format}}$ and $R_{\text{alignment}}$ provided in Section \ref{sec:experiments}.

\paragraph{Prompt Generator} is a language model designed to refine a given base prompt. First, it generates a reasoning trace about the refinement process, followed by the production of the final prompt (an example is shown in Appendix~\ref{sec:prompts_format}). 
At each training step, the Prompt Generator produces a set of outputs $o_1, \dots, o_n$ from which candidate prompts $p_1, \dots, p_n$ are extracted. These prompts are then passed to the Evaluation Model, which generates answers conditioned on each prompt.
We denote the Prompt Generator as $\pi_{\theta}^{\text{generator}}$.

\begin{table*}[ht]
\caption{Accuracy on classification tasks, averaged over three runs.
Colours mark the best (\redc{red}), second-best (\orangec{orange}) and third-best (\yellowc{yellow}) numbers in each column; minor differences ($\le{}0.05$) are treated as ties.
The right-most column shows the mean accuracy of each method across the seven datasets.}
\centering
\small
\renewcommand{\arraystretch}{1.2}
\resizebox{1.0\textwidth}{!}{%
\begin{tabular}{cccccccc:c}
\hline
\textbf{Method / Dataset} & \textbf{SST-2} & \textbf{CR} & \textbf{MR} & \textbf{SST-5} & \textbf{AG's News} & \textbf{TREC} & \textbf{Subj} & \textbf{Avg} \\ 
\hline
MI
& 92.70 & 87.25 & 87.40 & 52.31 & 82.29 & 69.20 & 57.95 & 75.59 \\

NI
& \yellowc{95.77} & 91.50 & \orangec{90.85} & 51.90 & 83.43 & 66.60 & 68.10 & 78.31 \\

APO
& 93.71\textsubscript{\textcolor{gray}{±0.25}}
& \redc{93.48\textsubscript{\textcolor{gray}{±0.24}}}
& 89.97\textsubscript{\textcolor{gray}{±1.37}}
& 53.94\textsubscript{\textcolor{gray}{±0.29}}
& \yellowc{83.73\textsubscript{\textcolor{gray}{±0.31}}}
& 71.30\textsubscript{\textcolor{gray}{±1.90}}
& 69.80\textsubscript{\textcolor{gray}{±5.96}}
& 79.42 \\

APE
& 91.23\textsubscript{\textcolor{gray}{±0.66}}
& \yellowc{92.87\textsubscript{\textcolor{gray}{±0.02}}}
& 89.90\textsubscript{\textcolor{gray}{±0.94}}
& 49.37\textsubscript{\textcolor{gray}{±5.66}}
& 82.58\textsubscript{\textcolor{gray}{±1.20}}
& \orangec{77.07\textsubscript{\textcolor{gray}{±1.61}}}
& 73.92\textsubscript{\textcolor{gray}{±1.39}}
& 79.56 \\

GA
& 94.65\textsubscript{\textcolor{gray}{±1.04}}
& 92.75\textsubscript{\textcolor{gray}{±0.40}}
& 90.45\textsubscript{\textcolor{gray}{±0.72}}
& 53.76\textsubscript{\textcolor{gray}{±1.13}}
& 82.24\textsubscript{\textcolor{gray}{±1.00}}
& \redc{79.20\textsubscript{\textcolor{gray}{±2.83}}}
& \orangec{74.93\textsubscript{\textcolor{gray}{±3.12}}}
& \orangec{81.14} \\

DE
& 93.29\textsubscript{\textcolor{gray}{±0.34}}
& \orangec{93.38\textsubscript{\textcolor{gray}{±0.19}}}
& 89.98\textsubscript{\textcolor{gray}{±0.24}}
& \orangec{55.25\textsubscript{\textcolor{gray}{±0.37}}}
& 82.18\textsubscript{\textcolor{gray}{±1.04}}
& \yellowc{76.47}\textsubscript{\textcolor{gray}{±0.38}}
& 73.08\textsubscript{\textcolor{gray}{±4.95}}
& \yellowc{80.52} \\

PromptAgent
& 94.58\textsubscript{\textcolor{gray}{±0.49}}
& 89.68\textsubscript{\textcolor{gray}{±1.27}}
& 88.62\textsubscript{\textcolor{gray}{±2.62}}
& 51.74\textsubscript{\textcolor{gray}{±1.02}}
& 83.44\textsubscript{\textcolor{gray}{±1.62}}
& 70.20\textsubscript{\textcolor{gray}{±1.27}}
& \orangec{76.83\textsubscript{\textcolor{gray}{±2.01}}}
& 79.30 \\

GRACE
& 93.61\textsubscript{\textcolor{gray}{±0.53}}
& 90.92 \textsubscript{\textcolor{gray}{±1.15}}
& 89.60\textsubscript{\textcolor{gray}{±1.51}}
& 53.96\textsubscript{\textcolor{gray}{±0.93}}
& 	82.34 \textsubscript{\textcolor{gray}{±0.39}}
& 72.53\textsubscript{\textcolor{gray}{±8.62}}
& 73.92\textsubscript{\textcolor{gray}{±3.05}}
& 79.55 \\

PromptWizard
& 93.87\textsubscript{\textcolor{gray}{±1.38}}
& 89.45\textsubscript{\textcolor{gray}{±0.25}}
& 89.30\textsubscript{\textcolor{gray}{±0.15}}
& 52.04\textsubscript{\textcolor{gray}{±0.14}}
& 82.82\textsubscript{\textcolor{gray}{±1.00}}
& 71.10\textsubscript{\textcolor{gray}{±2.10}}
& 73.05\textsubscript{\textcolor{gray}{±3.12}}
& 78.80 \\

\hline

\ours{} (–PS)
& \orangec{95.98\textsubscript{\textcolor{gray}{±0.19}}}
& 92.17\textsubscript{\textcolor{gray}{±0.02}}
& \yellowc{90.72\textsubscript{\textcolor{gray}{±0.05}}}
& \yellowc{54.80\textsubscript{\textcolor{gray}{±1.10}}}
& \orangec{83.84\textsubscript{\textcolor{gray}{±0.33}}}
& 72.00\textsubscript{\textcolor{gray}{±0.86}}
& 66.98\textsubscript{\textcolor{gray}{±2.86}}
& 79.50 \\

\ours{}
& \redc{96.32\textsubscript{\textcolor{gray}{±0.04}}}
& \yellowc{92.83\textsubscript{\textcolor{gray}{±0.24}}}
& \redc{91.27\textsubscript{\textcolor{gray}{±0.05}}}
& \redc{56.21\textsubscript{\textcolor{gray}{±0.15}}}
& \redc{84.36\textsubscript{\textcolor{gray}{±0.08}}}
& \orangec{77.07\textsubscript{\textcolor{gray}{±2.36}}}
& \redc{76.90\textsubscript{\textcolor{gray}{±0.95}}}
& \redc{82.14} \\
\hline
\end{tabular}%
}
\label{tab:cls}
\end{table*}

\paragraph{Evaluation Model.}
assesses the quality of each prompt generated by the Prompt Generator by evaluating its performance on a randomly sampled subset of the training data. For each prompt, a reward is computed for every observation based on its effectiveness, and these rewards are averaged to obtain a final score for the prompt.
The Evaluation Model is implemented as a frozen language model, used exclusively for inference.
We denote the Evaluation Model as $\pi^{\text{eval}}$.

\paragraph{Optimization}
After obtaining a reward for each prompt, we optimize the Prompt Generator using the Group Relative Policy Optimization (GRPO) update rule \citep{shao2024deepseekmath}. The illustration of this process can be found in Figure \ref{fig:training}. 

\paragraph{Prompt Selection}
As the prompt generator evolves during training and each version generates multiple prompts, we obtain a large selection of candidate prompts. 
We sample in a regular interval a number of prompts and test them on the validation set and keep the overall best one according to the task metric used.
The full algorithm is showcased in the Appendix~\ref{sec:deepprompt-pseudo-code}.

\section{Experiments}
\label{sec:experiments}

We evaluate PRL on classification, summarization, simplification, domain-knowledge, and mathematical reasoning tasks. As Prompt Generator and (frozen) Evaluation Model we use Qwen2.5-7B-Instruct~\citep{yang2024qwen2}, training a separate Prompt Generator for each task and dataset. Unless otherwise specified, all experiments run for 48 hours on two NVIDIA A100 GPUs (40\,GB each). To ensure fair comparison during Prompt Selection, we adopt the same scoring function as EvoPrompt and use an identical validation dataset for selection. Optimization, sampling/selection, and reward hyperparameters are given in Appendix~\ref{app:opt_details}, $r_{\text{format}}$ and $r_{\text{alignment}}$ are task-specific and defined below. We additionally provide benchmarks reproduction details in Appendix~\ref{sec:appendix_reproduction}.

\begin{table*}[ht]
\caption{Example of evolution of the best prompt over training iterations by \ours{}. Notably, although few-shot prompting is never explicitly encouraged by the objective, the model spontaneously begins to insert synthetic few-shot examples as training progresses. This emergent behavior is consistently observed across other benchmarks.}
\label{tab:prompt_evolution}
\centering

\begin{tcolorbox}[
  enhanced,
  colback=white,
  colframe=blue!65!black,
  boxrule=0.6pt,
  arc=3pt,
  left=6pt,right=6pt,top=6pt,bottom=6pt,
]
{\small
\setlength{\tabcolsep}{6pt}
\renewcommand{\arraystretch}{1.22}

\rowcolors{2}{rowgray}{white}

\begin{tabularx}{\linewidth}{>{\centering\arraybackslash\bfseries}m{1.6cm} >{\raggedright\arraybackslash}X}
\toprule
\rowcolor{rowgray}
\textbf{Iteration} & \textbf{Prompt (abridged)} \\
\midrule

\cellcolor{blue!6}100  &
Classify the sentiment of the given movie-review sentence as \texttt{positive} or \texttt{negative}. \textit{[...]} Return only the label. \\
\cmidrule(lr){1-2}

\cellcolor{blue!6}500  &
Classify each movie-review sentence as \texttt{positive} or \texttt{negative} based on overall sentiment. \textit{[...]} Output only \texttt{positive} or \texttt{negative}. \\
\cmidrule(lr){1-2}

\cellcolor{blue!6}1200 &
Classify each movie-review sentence as \texttt{positive} or \texttt{negative}, considering context/nuance. \textbf{Examples:}\newline
\quad ``The acting was superb, and the plot was engaging.'' $\rightarrow$ \texttt{positive}\newline
\quad ``The movie was so slow and boring that I almost fell asleep.'' $\rightarrow$ \texttt{negative}\newline
Return only \texttt{positive} or \texttt{negative}. \\
\bottomrule
\end{tabularx}
}
\end{tcolorbox}
\end{table*}

\subsection{Baseline Methods}\label{par:benchmarks}
We benchmark \ours{} against both human-written task-specific prompts and a range of general-purpose prompt engineering algorithms.

\begin{itemize}
    \item \texttt{MI (Manual Instruction)} \citep{zhang2022opt} and NI (Natural Instruction) \citep{mishra2021cross}: Human-written task instructions used as prompts.

    \item \texttt{APE (Automatic Prompt Engineer)} \citep{zhou2022large}: Generates multiple candidate prompts with an LLM and selects the best via performance-based filtering.

    \item \texttt{APO (Automatic Prompt Optimization)} \citep{pryzant2023automatic}: Iteratively refines prompts using natural-language critiques and search (e.g., beam search), without gradients.

    \item \texttt{EvoPrompt} \citep{guo12connecting}: Evolves a population of prompts with evolutionary operators (selection, mutation, crossover) guided by task performance.
    \begin{itemize}
        \item \texttt{DE (Differential Evolution)}: Uses differential-evolution style updates to propose and mix prompt candidates.
        \item \texttt{GA (Genetic Algorithm)}: Uses standard genetic-algorithm selection and crossover/mutation to evolve prompts.
    \end{itemize}
    
    \item \texttt{PromptAgent} \citep{wang2023promptagent}: An LLM  that iteratively edits prompts using automatic feedback from evaluations on held-out data.

    \item \texttt{GRACE} \citep{shi2025lossgaingatedrefinement}: Refines prompts with a gated two-stage update rule and compression to keep only changes that improve validation performance. 
    
    \item \texttt{PromptWizard} \cite{agarwal2025promptwizard}: Iteratively critiques and rewrites prompts using task feedback, balancing exploration and exploitation. It can also synthesize novel in-context examples rather than relying solely on demonstrations drawn from the training set.

\end{itemize}
\renewcommand{\arraystretch}{1.3}

\noindent
We present a comparison of various methods in Figure ~\ref{fig:cycle_and_comparison} (right).
To ensure a fair comparison, we utilize the Qwen2.5-7B-Instruct model across all methods,
serving both as the prompt generator and the Evaluation Model. 
Due to substantial differences between PRL and RLPrompt~\citep{deng2022rlprompt} that prevent a fair comparison, we exclude RLPrompt from the main experiments. We instead conduct a dedicated comparative study in Appendix~\ref{sec:appendix_reproduction}.
\subsection{Results} 
\paragraph{Classification}
For classification, we evaluate on (i) binary sentiment datasets SST-2 \citep{socher2013recursive}, MR \citep{pang2005seeing}, and CR \citep{hu2004mining}, (ii) multiclass sentiment SST-5 \citep{socher2013recursive}, (iii) question type classification TREC \citep{voorhees2000building}, (iv) news topic classification AG's News \citep{zhang2015character}, and (v) subjectivity classification SUBJ \citep{pang2004sentimental}).

We apply a unified reward function across all classification tasks, with reward parameters set as \( r_{\text{format}} = r_{\text{alignment}} = 1 \). The component \( r_{\text{format}} \) is specifically awarded when the Evaluation Model's output is a valid label, i.e.\ one that belongs to the task's set of permissible labels. The scoring function \( f \) used in all classification tasks is accuracy. We report classification results in Table~\ref{tab:cls} (additional details in Appendix~\ref{app:opt_details}). 
Overall, \ours{} achieves state-of-the-art average performance across the classification benchmarks. 
Interestingly, we consistently observe an emergent behavior during training: as optimization progresses, the prompt generator begins to include few-shot examples even though this is not explicitly incentivized by the objective. 
An illustrative example is shown in Table~\ref{tab:prompt_evolution}, and we observe the same phenomenon across most other tasks. We provide a qualitative prompt comparison between PRL, Manual Instruction, and EvoPrompt on SST-2 in Appendix~\ref{app:prompt_comparison}. The remaining prompts for the other classification tasks are reported in Appendix~\ref{sec:prompts_cls}.

\paragraph{Summarization}  
We evaluate \ours{} on a summarization task, where the model is required to extract and condense the most important information from a given text. Our experiments are conducted on the \textsc{SAMSum} dataset \citep{gliwa2019samsum}, which consists of English chat dialogues resembling real-life messenger conversations. We evaluate summarization quality using ROUGE \citep{lin2004rouge}, reporting ROUGE-1 (unigram overlap), ROUGE-2 (bigram overlap), and ROUGE-L (longest common subsequence) with respect to the reference summaries. For this task, we set \( r_{\text{format}} = 0 \), as summarization does not involve selecting from a fixed label set.
Instead, we use \( r_{\text{alignment}} \), which computes the reward based on the average of the three ROUGE metrics. The results, shown in Table~\ref{tab:sum}, indicate that \ours{} significantly outperforms all baseline methods on the summarization task except PromptAgent. Interestingly, \ours{} consistently opts to generate prompts without incorporating few-shot examples. We provide a qualitative comparison of prompts from Manual Instruction, EvoPrompt, and \ours{} in Appendix~\ref{app:prompt_comparison}, together with the corresponding average ROUGE scores.
Although the two EvoPrompt prompts are nearly identical, they produce markedly different results, highlighting the sensitivity of LLMs to small phrasing changes.

\begin{table}
\caption{Text summarization results.}
  \centering
  \small
  {\renewcommand{\arraystretch}{1.3}%
  \resizebox{\columnwidth}{!}{%
  \begin{tabular}{c|ccc}
    \hline
    \textbf{Method} & \textbf{ROUGE-1} & \textbf{ROUGE-2} & \textbf{ROUGE-L} \\
    \hline
    MI  & 32.76 & 10.39 & 28.97 \\
    APE 
        & 37.12\textsubscript{\textcolor{gray}{±2.02}} 
        & 12.97\textsubscript{\textcolor{gray}{±0.74}} 
        & 33.32\textsubscript{\textcolor{gray}{±1.68}} \\
    GA  
        & 39.69\textsubscript{\textcolor{gray}{±1.76}} 
        & 14.47\textsubscript{\textcolor{gray}{±1.00}} 
        & 35.84\textsubscript{\textcolor{gray}{±1.63}} \\
    DE  
        & 33.91\textsubscript{\textcolor{gray}{±4.04}} 
        & 12.53\textsubscript{\textcolor{gray}{±1.47}} 
        & 31.05\textsubscript{\textcolor{gray}{±3.79}} \\

    PromptAgent  
        & \redc{44.59\textsubscript{\textcolor{gray}{±1.10}}}
        & \redc{17.13\textsubscript{\textcolor{gray}{±1.54}}}
        & \redc{38.09\textsubscript{\textcolor{gray}{±1.26}}}\\

    GRACE  
        & \yellowc{40.61}\textsubscript{\textcolor{gray}{±0.54}} 
        & \yellowc{14.65}\textsubscript{\textcolor{gray}{±0.53}} 
        & \yellowc{35.86}\textsubscript{\textcolor{gray}{±0.54}} \\

    \ours{}  
        & \orangec{42.47}\textsubscript{\textcolor{gray}{±0.83}} 
        & \orangec{16.17}\textsubscript{\textcolor{gray}{±0.24}} 
        & \orangec{37.73}\textsubscript{\textcolor{gray}{±0.36}} \\
    \hline
  \end{tabular}%
  }}%
  \label{tab:sum}
\end{table}

\paragraph{Simplification}
We evaluate \ours{} on sentence simplification using the \textsc{ASSET} dataset \citep{alva2020asset}, a crowdsourced multi-reference benchmark covering diverse rewriting operations (e.g., paraphrasing, splitting, deletion, and reordering). We evaluate simplification with SARI \citep{xu2016optimizing}, which compares the output to the source and reference simplifications by scoring added, deleted, and kept n-grams. We set \(r_{\text{format}}=0\) and use SARI for both the alignment reward and the final evaluation score. The results are presented in Table~\ref{tab:subj_robustness} (left). 
Our baselines perform on average comparable to a manually written prompt. Generated prompts from \ours{}, EvoPrompt, and Manual Instruction are provided in Appendix ~\ref{app:prompt_comparison}.
For this task baselines generated comparatively simple prompts which fail to provide sufficient guide the model's output.
In contrast, the \ours\  prompt is precise, comprehensive, includes a well-constructed example and leads to a substantial performance improvement.

\paragraph{Domain Knowledge Task}
To assess \ours{} on a domain-knowledge benchmark, we evaluate it on MedQA \citep{yang2025llmmedqaenhancingmedicalquestion}, a multiple-choice medical QA dataset with answer options (A--E). We assign a format reward when the model outputs a valid option letter, and use accuracy as the alignment reward. Results are reported in Table~\ref{tab:math_tasks}. \ours{} achieves the best performance among the compared methods, indicating that it also yields gains on tasks that require specialized domain knowledge.

\paragraph{Reasoning Tasks}\label{par:math_tasks}
We further evaluate \ours{} on mathematical reasoning benchmarks, including GSM8K \citep{cobbe2021gsm8k}, MATH500 \citep{lightman2023lets}, and DeepMath \citep{deepmath}. For these tasks, we use the same reward formulation as in classification. As shown in Table~\ref{tab:math_tasks}, \ours{} consistently outperforms the baselines, achieving state-of-the-art results and indicating that PRL can optimize prompts for multi-step reasoning beyond standard classification and generation settings. We additionally report an experiment on GSM8K in Appendix~\ref{sec:appendix_reproduction} under a stricter evaluation protocol that considers a prediction correct only if it outputs the final integer answer, without any accompanying reasoning trace.


\begin{table}[t]
  \caption{Math reasoning and domain-specific task results.}
  \label{tab:math_tasks}
  \centering

  {\renewcommand{\arraystretch}{1.3}%
  \resizebox{\columnwidth}{!}{%
  \begin{tabular}{ccccc}
    \toprule
    \textbf{Method} 
    & \textbf{GSM8K} 
    & \textbf{MATH500} 
    & \textbf{DeepMath} 
    & \textbf{MedQA} \\
    \midrule
    APE    
      & \orangec{83.43}\textsubscript{\textcolor{gray}{±1.98}} 
      & 31.53\textsubscript{\textcolor{gray}{±1.04}} 
      & 15.47\textsubscript{\textcolor{gray}{±0.45}} 
      & 45.66\textsubscript{\textcolor{gray}{±0.97}} \\
    DE     
      & 79.52\textsubscript{\textcolor{gray}{±0.45}} 
      & 34.20\textsubscript{\textcolor{gray}{±1.39}}
      & 16.10\textsubscript{\textcolor{gray}{±0.00}} 
      & 51.76\textsubscript{\textcolor{gray}{±0.16}} \\
    GA     
      & 81.62\textsubscript{\textcolor{gray}{±1.38}} 
      & \orangec{40.13}\textsubscript{\textcolor{gray}{±1.39}} 
      & \orangec{18.63}\textsubscript{\textcolor{gray}{±2.37}} 
      & \yellowc{51.95}\textsubscript{\textcolor{gray}{±1.61}} \\
      
         PromptAgent  
        & 78.20\textsubscript{\textcolor{gray}{±0.00}} 
        & \yellowc{36.20\textsubscript{\textcolor{gray}{±1.80}}}
        &16.43\textsubscript{\textcolor{gray}{±0.96}}  & 51.87\textsubscript{\textcolor{gray}{±0.96}}\\
        
    GRACE  
      & 82.37\textsubscript{\textcolor{gray}{±1.82}} 
      & 33.20\textsubscript{\textcolor{gray}{±1.60}} 
      & 15.05\textsubscript{\textcolor{gray}{±0.16}} 
      & \orangec{52.26}\textsubscript{\textcolor{gray}{±0.16}} \\
    
    PromptWizard  
      & \redc{86.61}\textsubscript{\textcolor{gray}{±0.19}} 
      & 35.50\textsubscript{\textcolor{gray}{±0.30}} 
      & \yellowc{18.52\textsubscript{\textcolor{gray}{±1.08}}}
      & 51.84\textsubscript{\textcolor{gray}{±0.23}}\\
      
    PRL    
      & \orangec{86.15}\textsubscript{\textcolor{gray}{±0.55}} 
      & \redc{44.40}\textsubscript{\textcolor{gray}{±1.40}} 
      & \redc{21.58}\textsubscript{\textcolor{gray}{±0.22}} 
      & \redc{53.34}\textsubscript{\textcolor{gray}{±0.11}} \\
    \bottomrule
  \end{tabular}%
  }}%

  \vskip -0.1in
\end{table}

\subsection{Ablations}

\paragraph{Ablation Study: Influence of Prompt Selection} 
We analyze the impact of the Prompt Selection process on overall performance.
In the ablation setting, instead of selecting the best prompt iteratively during training, we simply report the final prompt at the end of training. To do so after training we sample \( n_{\text{test}} \) prompts to choose the best one according to the validation set. This comparison is performed across all classification tasks. The results, shown in Table~\ref{tab:cls}, demonstrate that Prompt Sampling not only improves final performance but also enhances training efficiency. By selecting strong prompts throughout the training process, Prompt Selection leads to both better and faster results. 

\paragraph{Ablation Study: Influence of Reasoning}

To investigate the role of explicit reasoning in our method, we conduct an ablation study based on the prompt design illustrated in Appendix~\ref{sec:prompts_format}. In the standard setup, the model is instructed to perform a reasoning process before producing the final answer. To evaluate the effect of removing this step, we modify the prompt to omit the reasoning phase and instead directly request the model to generate the answer within \texttt{<answer>} \texttt{</answer>} tokens. We perform this experiment on the SUBJ dataset, training two identical models, except for either using or or omitting reasoning.
This difference leads to a substantial drop in accuracy, from 75.05 (±1.63) with reasoning to 60.12 (±1.62) without reasoning, showing the importance of explicit reasoning in our approach.

\paragraph{Ablation Study: \ours{} on Larger Models}

To study how \ours{} scales with stronger evaluation models, we use Qwen2.5-\{7B, 14B, 32B\}-Instruct as the Evaluation Model while keeping Qwen2.5-7B-Instruct as the Prompt Generator.

\begin{table}[h]
\caption{Comparison of accuracy across different model sizes of the Evaluation Model on the MR dataset.}
\centering
\small
\renewcommand{\arraystretch}{1.2}
\resizebox{\columnwidth}{!}{%
\begin{tabular}{c|ccc}
\hline
\textbf{Parameters} & \textbf{7B} & \textbf{14B} & \textbf{32B} \\
\hline
MI      & 87.40 & 89.20 & 90.15 \\
\ours{} & 91.27\textsubscript{\textcolor{gray}{±0.05}} & 92.03\textsubscript{\textcolor{gray}{±0.13}} & 92.52\textsubscript{\textcolor{gray}{±0.02}} \\
\hline
\end{tabular}%
}
\label{tab:abl_size}
\end{table}

\noindent
We compare the performance using the base prompt against prompts generated by \ours{} on the MR dataset. The results, presented in Table~\ref{tab:abl_size}, show that all model sizes benefit from \ours{}, demonstrating two key findings:
    (i)~Even larger LLMs remain vulnerable to prompt variation.
    (ii)~\ours{} is capable of effectively tailoring prompts for both smaller and larger models, significantly improving their performance.

\paragraph{Ablation Study: PRL Beyond Qwen} 
We test the cross model robustness of \ours{} using LLaMA 3.1-8B- Instruct~\cite{llama3modelcard} on summarization: 
\begin{itemize}
    \item First, we assess prompt portability: prompts learned by \ours{} and benchmark methods with Qwen2.5-7B- Instruct (as both prompt generator and evaluator during training) are applied unchanged to LLaMA-3.1-8B Instruct . As shown in Table~\ref{tab:llama}, these prompts transfer well and remain competitive with strong baselines, indicating that \ours{} produces prompts that generalize beyond the backbone used to train them.
    \item Second, motivated by observations of potential spurious gains when training with Qwen under weak reward signals~\cite{shao2025spurious}, we retrain \ours{} using LLaMA as the prompt generator and Qwen as the prompt evaluator (for both training and test). This variant is reported as PRL-LLaMA in Table~\ref{tab:llama}. The resulting prompts achieve equal or better ROUGE 1/2 than the Qwen trained counterpart.
    \item Third, to further assess the generality of prompts produced by \ours{} across evaluation models, we crafted the experiment in which we train \ours{} and benchmark methods with Qwen as prompt generator and Llama as evaluator (for both training and test). The results are presented in Table~\ref{tab:qwen_gen_llama_eval}. 
\end{itemize}
\noindent
Together, these findings suggest that \ours{}’s improvements are not an artifact of a particular model family: PRL trained prompts both transfer across architectures and train effectively on alternative backbones.

\begin{table}[h]
\caption{Cross-model generalization on summarization}
\label{tab:llama}
\centering
\small
{\renewcommand{\arraystretch}{1.2}%
\resizebox{\columnwidth}{!}{%
\begin{tabular}{lccc}
\toprule
\textbf{Method} 
& \textbf{ROUGE-1} & \textbf{ROUGE-2} & \textbf{ROUGE-L} \\
\midrule
MI & 33.33 & 10.77 & 27.12 \\ 
APE & 34.12\textsubscript{\textcolor{gray}{±3.86}} & \yellowc{12.90}\textsubscript{\textcolor{gray}{±2.56}} & \yellowc{25.72}\textsubscript{\textcolor{gray}{±3.63}} \\
GA & \orangec{38.73}\textsubscript{\textcolor{gray}{±2.36}} & \orangec{14.69}\textsubscript{\textcolor{gray}{±1.01}} & \orangec{30.38}\textsubscript{\textcolor{gray}{±2.94}} \\
DE & \yellowc{34.25}\textsubscript{\textcolor{gray}{±3.56}} & 11.57\textsubscript{\textcolor{gray}{±2.86}} & 25.43\textsubscript{\textcolor{gray}{±3.65}} \\
PRL & \redc{39.38}\textsubscript{\textcolor{gray}{±2.82}} & \redc{15.77}\textsubscript{\textcolor{gray}{±1.56}} & \redc{30.57}\textsubscript{\textcolor{gray}{±2.80}} \\ 
\midrule
PRL-LLaMA  & 43.70\textsubscript{\textcolor{gray}{±0.02}} & 16.66\textsubscript{\textcolor{gray}{±0.25}} & 37.46\textsubscript{\textcolor{gray}{±0.5}} \\
\bottomrule
\end{tabular}%
}}%
\end{table}

\begin{table}[t]
\caption{Summarization results using Qwen as prompt generator and LLaMA as evaluator.}
\label{tab:qwen_gen_llama_eval}
\centering
\small
{\renewcommand{\arraystretch}{1.2}%
\resizebox{\columnwidth}{!}{%
\begin{tabular}{lccc}
\toprule
\textbf{Method} & \textbf{ROUGE-1} & \textbf{ROUGE-2} & \textbf{ROUGE-L} \\
\midrule
MI   
& 28.92 
& 9.29 
& 25.09 \\
APE  
& \orangec{38.47}\textsubscript{\textcolor{gray}{±0.66}} 
& \yellowc{12.96}\textsubscript{\textcolor{gray}{±0.14}} 
& \yellowc{33.17}\textsubscript{\textcolor{gray}{±0.22}} \\
GA   
& \yellowc{37.48}\textsubscript{\textcolor{gray}{±2.65}} 
& \orangec{13.92}\textsubscript{\textcolor{gray}{±1.14}} 
& \orangec{33.48}\textsubscript{\textcolor{gray}{±2.54}} \\
PRL  
& \redc{40.06}\textsubscript{\textcolor{gray}{±0.02}} 
& \redc{13.98}\textsubscript{\textcolor{gray}{±0.12}} 
& \redc{34.06}\textsubscript{\textcolor{gray}{±0.01}} \\
\bottomrule
\end{tabular}%
}}%
\end{table}

\begin{figure*}[ht]
\begin{subfigure}[t]{0.65\textwidth}
  \begin{minipage}[t]{0.49\textwidth}
    \centering
    \includegraphics[width=\linewidth]{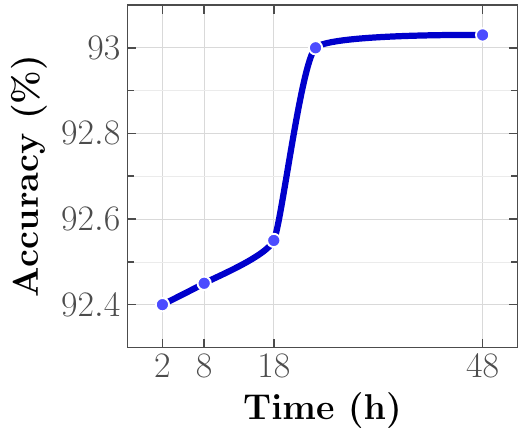}
  \end{minipage}
  \begin{minipage}[t]{0.49\textwidth}
    \centering
    \includegraphics[width=\linewidth]{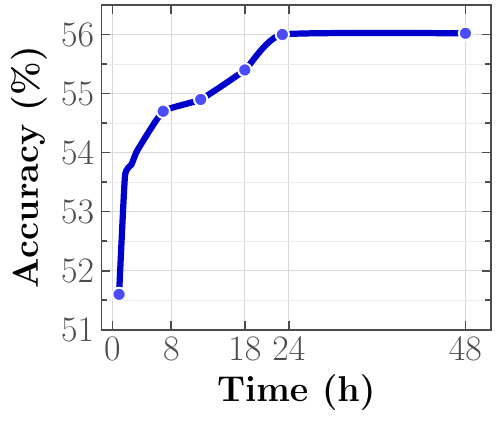}
  \end{minipage}
\caption{\underline{Left}: Accuracy over time for CR (left) and SST-5 (right). Although \ours{} requires more compute, strong prompts emerge well before the end of training and can be deployed early.}
  \label{fig:ablation_time}
\end{subfigure}
\hfill
\begin{subfigure}[t]{0.32\textwidth}
\centering
  \includegraphics[width=0.95\linewidth]{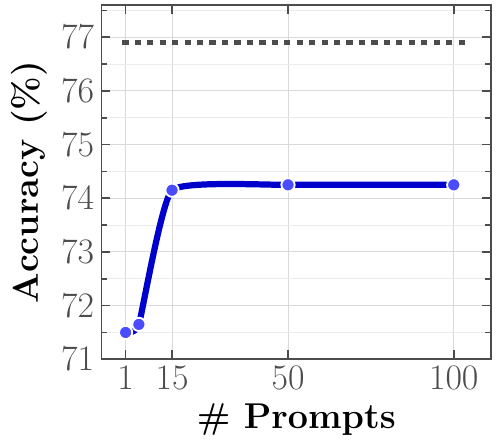}
\caption{SUBJ test accuracy vs.\ number of prompts sampled from Qwen2.5-72B-Instruct. The dashed line is average \ours{} performance. Unguided sampling saturates quickly without optimization.}
  \label{fig:ablation_qwen72b}
\end{subfigure}
\end{figure*}

\paragraph{Ablation Study: Sensitivity to the Base Prompt}
To assess the dependence of PRL on the initial base prompt, we reran the SUBJ experiment using a previously optimized PRL prompt as the starting point instead of a manual instruction. PRL converges to nearly identical performance and produces structurally similar prompts, indicating that the method is not overly sensitive to the exact wording of the base prompt as long as it specifies the task and label space. We further observe that iteratively replacing the base prompt during training does not yield additional gains, suggesting that a fixed minimal task description is sufficient in practice. We showcase optimized prompts and Accuracy results in Appendix \ref{app:prl_on_better_prompt}.

\paragraph{Ablation Study: Computational Cost}

\ours{} is more compute-intensive than some prior methods. However, its compute cost is front-loaded and amortized. Although we report a 48-hour budget, the built-in Prompt Selection mechanism surfaces strong prompts well before the end of training, these intermediate prompts can be deployed immediately. We illustrate this with the evolution of the final prompts' performance over time for CR and SST-5 in Figure \ref{fig:ablation_time}. The higher compute cost of \ours{} is a natural consequence of its deliberately larger search space: whereas APO, EvoPrompt and GRACE mainly operate by rephrasing a base prompt via critique loops or evolutionary edits (thus constraining exploration), PRL does not impose such restrictions. This broader search is precisely what enables PRL to synthesize novel few-shot examples and incorporate explicit reasoning structures, capabilities that, in turn, drive the observed performance gains.

\begin{table}[ht]
\small
\centering
\vspace{-0.6em}
\caption{\underline{Left}: Simplification results. \underline{Right}: \ours{} results on SUBJ when training with a smaller prompt generator.}
\begin{minipage}[t]{0.45\columnwidth}
\centering
\begin{adjustbox}{max width=\linewidth}
\label{tab:sim}
    \begin{tabular}{cc}
        \hline
        \textbf{Method} & \textbf{SARI} \\ \hline
        MI     & 43.77 \\

    APE    & 45.33\textsubscript{\textcolor{gray}{±0.83}} \\

    GA     & \yellowc{46.25}\textsubscript{\textcolor{gray}{±0.47}} \\

    DE     & 45.79\textsubscript{\textcolor{gray}{±0.35}} \\
    
        PromptAgent     & 45.12 \textsubscript{\textcolor{gray}{±0.57}}  \\

    GRACE  & \orangec{50.21}\textsubscript{\textcolor{gray}{±0.18}} \\

    \ours{} & \redc{52.26}\textsubscript{\textcolor{gray}{±3.51}} \\
        \hline
    \end{tabular}
\end{adjustbox}
\end{minipage}
\hfill
\begin{minipage}[t]{0.5\columnwidth}
\centering
\begin{adjustbox}{max width=\linewidth}
\begin{tabular}{lc}
\toprule
\textbf{Generator} & \textbf{Acc.} \\
\midrule
3B-Instr. & 74.62\textsubscript{\textcolor{gray}{±0.83}} \\
7B-Instr. & 76.90\textsubscript{\textcolor{gray}{±0.95}} \\
\bottomrule
\end{tabular}
\end{adjustbox}
\end{minipage}

\label{tab:subj_robustness}
\vspace{-0.6em}
\end{table}

\paragraph{Ablation Study: Motivation for a Dedicated Prompt Generator}
One might expect that a sufficiently strong LLM could produce high-quality prompts via unguided sampling. In practice, this is not enough. Even when we sample up to 100 prompt candidates from a much larger generator (Qwen2.5-72B-Instruct), select the best on the validation set, and evaluate it on the test set, performance saturates quickly and remains well below \ours{}, as shown in Figure~\ref{fig:ablation_qwen72b}. This indicates that the prompt space is highly non-trivial: guided RL-based optimization explores it more effectively than brute-force sampling and yields consistently stronger prompts.

\paragraph{Ablation Study: Prompt Generator Size}
To investigate, whether \ours{} is capable of generating good prompts with smaller generator, we run the experiemnts where we change the generator to Qwen2.5-3B-Instruct.
To investigate how does \ours{} behave w The obtained results presented in Table~\ref{tab:subj_robustness} (right) show that PRL works with a smaller 3B generator and that performance improves as the generator becomes larger, as one would expect.

 \noindent
We perform additional ablation studies in Appendix \ref{sec:abl_appendix}.

\section{Conclusions}
We introduced \ours{}, an RL-based method for prompt generation that consistently outperforms prior approaches across classification, summarization, and simplification tasks. 
Our results show that, even with recent instruction-tuned LLMs, prompt choice can substantially affect performance, indicating persistent sensitivity to semantically equivalent variations, including for the largest evaluation model we tested (Qwen2-32B-Instruct). 
We also find that effective prompting is task-dependent: some tasks benefit from few-shot demonstrations or specific semantic cues, while others do not. 
By jointly optimizing instructions and (when useful) few-shot content, \ours{} navigates these trade-offs and produces strong task-specific prompts.

\newpage

\section{Limitations}
While \ours{} achieves strong results across diverse benchmarks, we note several limitations.

\begin{description}[leftmargin=!, labelwidth=10pt]
\item[Need for labeled data:]
\ours{} optimizes prompts using task-level rewards (e.g., accuracy, ROUGE, SARI) and therefore requires supervision to score candidate prompts reliably. This may limit direct applicability in fully unsupervised settings.

\item[Compute overhead:]
Training the prompt generator with RL is more compute-intensive than lighter-weight search or critique-based baselines. In practice, however, strong prompts often emerge well before the full training budget (via Prompt Selection), and the resulting prompts tend to transfer across evaluation models, partially amortizing this cost.

\item[Task-specific training:]
In our current setup, we train a separate prompt generator per task/dataset. Developing a more universal prompt generator that transfers across tasks with minimal adaptation is an important direction for future work.

\item[Prompt length and inference cost:]
When \ours{} chooses to include few-shot demonstrations or richer instructions, prompts can become longer, increasing test-time token usage and latency. 

\item[Scaling beyond 32B evaluation models:]
We evaluate \ours{} with evaluation models up to 32B parameters. While gains are consistent across these sizes, further study is needed to characterize performance and efficiency for substantially larger or frontier-scale models.
\end{description}

\section*{Ethics Statement}
We conducted this research in line with the ACL Code of Ethics and the ACM Code of Ethics and Professional Conduct. We note that large language models can be misapplied, regardless of whether prompts are engineered manually or automatically, to produce deceptive, offensive, or otherwise harmful outputs. Our goal in developing automated prompt optimization is to improve controllability and transparency of model behavior, thereby supporting safer and more responsible deployment. At the same time, effective real-world use requires continued risk assessment, robust evaluation, and appropriate safeguards to reduce potential harms.

\bibliography{custom}

\appendix

\newpage
\section{Appendix - \ours{} Pseudo Code}
\label{sec:deepprompt-pseudo-code}
In Algorithm \ref{alg} we provide the \ours{} algorithm.
\begin{algorithm}
\caption{
\ours{} 
}
\label{alg:prompt_training}
\begin{algorithmic}[1]
\REQUIRE 
    $\pi_{\theta}^{\text{generator}}$: prompt generator \\
    $\pi^{\text{eval}}$: frozen Evaluation Model \\
    $T, V$: training and validation datasets \\
    $n, n_{\text{test}}$: number of prompts during training/Prompt Selection \\
    $k$: number of samples per iteration \\
    $I$: total number of iterations \\
    $t$: Prompt Selection frequency \\
    $R$: reward operator \\
    $f$: scoring function 
\vspace{0.5em}
\STATE $best\_score \gets 0$, $best\_prompt \gets \mathtt{``"}$
\FOR{$i = 1$ to $I$}
    \STATE Sample $k$ training samples $D \sim T$
    \STATE Generate answers $o_1, \dots, o_n \sim \pi_\theta^{\text{generator}}$
    \STATE Extract answers $p_1, \dots, p_n$ from  $o_1, \dots, o_n$
    \STATE Compute rewards $r_j = R(\pi^{\text{eval}}, D, p_j, o_j)$ for each $j=1,...,n$
    \STATE Update $\theta$ using GRPO with rewards $\{r_j\}$
    \IF{$i \bmod t = 0$}
        \STATE Generate test prompts $p_1, \dots, p_{n_{\text{test}}} \sim \pi_\theta^{\text{generator}}$
        \STATE \textbf{with} \texttt{torch.no\_grad()}:
        \STATE Compute scores $s_j = f(\pi^{\text{eval}}, V, p_j)$
        \STATE Let $j^* = \arg\max_j s_j$
        \IF{$s_{j^*} > best\_score$}
            \STATE $best\_score \gets s_{j^*}$, $best\_prompt \gets p_{j^*}$
        \ENDIF
    \ENDIF
\ENDFOR 
\RETURN $best\_prompt$
\end{algorithmic}
\label{alg}
\end{algorithm}

\section{Appendix - PRL Prompt Format}
\label{sec:prompts_format}
In Figure \ref{fig:prompt} we present the prompt format used by PRL. In the system prompt, we instruct the model to generate a reasoning trace enclosed within <think> and </think> tokens, followed by the final answer encapsulated within <answer> and </answer> tokens. The user message provides the base prompt that it should refine. The model’s objective is to produce the prompt that is better than the best prompt.

\begin{figure*}[htbp]
\tcbset{
  tightsplit/.style={
    enhanced,
    before upper=\vspace{-8pt},
    after upper=\vspace{-6pt},
    before lower=\vspace{-8pt},
    after lower=\vspace{-8pt},
    boxsep=1pt, left=4pt, right=4pt, top=10pt, bottom=10pt,
    center title,
    fonttitle=\scriptsize
  }
}
\tcbset{every box/.style={fonttitle=\tiny}}
\centering
\begin{minipage}[t]{0.98\textwidth}
  \tcbset{tightsplit,colback=blue!5,colframe=blue!70!black,title=System Prompt}
  
  \begin{tcolorbox}
    \scriptsize A conversation between User and Assistant. The user asks a question, and the Assistant solves it. The assistant first thinks about the reasoning process in the mind and then provides the user with the answer. The reasoning process and answer are enclosed within <think> </think> and <answer> </answer> tags, respectively, i.e., <think> reasoning process here </think><answer> answer here </answer>
  \end{tcolorbox}
\end{minipage}\hfill
\begin{minipage}[t]{0.98\textwidth}
  \tcbset{tightsplit,colback=green!5,colframe=green!70!black,title=User Prompt}
  \vspace{0pt}
  \begin{tcolorbox}
    \scriptsize Your task is to refine a base prompt for another model that performs a sentiment classification task. Improve the instructions to enhance the model's performance. The base prompt: 
    In this task, you are given sentences from movie reviews. The task is to classify a sentence as 'positive' if the sentiment of the sentence is positive or as 'negative' if the sentiment of the sentence is negative.
    Return label 'positive' or 'negative' only without any other text.
  \end{tcolorbox}
\end{minipage}
\caption{Prompt used by \ours{}}
\label{fig:prompt}
\end{figure*}

\section{Optimization, Sampling, and Reward Details}
\label{app:opt_details}
\paragraph{Optimization.}
We fine-tune using Group Relative Policy Optimization (GRPO)~\citep{zhao2024swiftascalablelightweightinfrastructure} with $\epsilon=0.2$, $\beta=0.04$, and weight decay $0.1$. We additionally apply Low-Rank Adaptation (LoRA)~\citep{hu2022lora} with learning rate $1\times10^{-6}$, $\alpha=32$, and rank $r=8$.

\paragraph{Sampling and Prompt Selection.}
During training, we sample $n=4$ prompts per iteration and perform Prompt Selection every 100 iterations.

\paragraph{Shared reward hyperparameters.}
Across all tasks we set $r_{\text{token}} = r_{\text{structure}} = 0.75$ unless stated otherwise. The rewards $r_{\text{format}}$ and $r_{\text{alignment}}$ are task-specific and defined in the corresponding experimental sections.

\paragraph{Classification}
For classification, we select the final prompt by sampling \(n_{\text{test}} = 10\) test prompts. 
During training-time evaluation, we use a subset of 100 training examples for most datasets, for CR and AG's News, we reduce this to 30 examples due to longer inputs and higher evaluation cost.

\section{Appendix - Prompt Comparison}
\label{app:prompt_comparison}

In this section, we compare prompts of \ours{} against Manual Instruction and EvoPrompt on classification, summarization and simplification tasks.

\begin{figure}[htbp]
\centering
\tcbset{
  tightsplit/.style={
    enhanced,
    before upper=\vspace{-8pt},  after upper=\vspace{-6pt},
    before lower=\vspace{-8pt},  after lower=\vspace{-8pt},
    boxsep=1pt, left=4pt, right=4pt, top=10pt, bottom=10pt,
    center title,  
    fonttitle=\scriptsize
  }
}

\newcommand{\promptbox}[4]{
  \begin{tcolorbox}[tightsplit, colback=#1!5, colframe=#1!75!black, title=#2]
    \scriptsize #3%
    \tcblower
    \centering\scriptsize Acc.: #4
  \end{tcolorbox}
}

\begin{minipage}[ht]{1.0\linewidth}\vspace{0pt}
  \promptbox{red}{\ours}{%
    In this task, you will classify the sentiment of movie review
    sentences as ‘positive’ or ‘negative’. Examples: \\
    ``The movie was thrilling and exciting’’ → positive; \\
    ``The plot was boring and predictable’’ → negative. \\
    Return only the label.}{96.38}
\end{minipage}
\hfill
\begin{minipage}[t]{0.49\linewidth}\vspace{0pt}
  \promptbox{cyan}{Manual Instruction}{%
    Please perform Sentiment Classification. Given the
    sentence, assign a label from [‘negative’, ‘positive’].
    Return the label only.}{92.70}
\end{minipage}
\hfill
\begin{minipage}[t]{0.49\linewidth}\vspace{0pt}
  \promptbox{orange}{EvoPrompt (GA)}{%
    Rephrase the movie-review snippet and assign a sentiment
    label from [‘positive’, ‘negative’]. Provfide only the label.}{95.83}
\end{minipage}

\caption{Comparison of a manual instruction, the best \ours{} prompt,
and the best EvoPrompt prompt along with their accuracies on SST-2 task.}
\label{fig:subj}
\end{figure}

\begin{figure}[htbp]
\tcbset{
  tightsplit/.style={
    enhanced,
    before upper=\vspace{-8pt},
    after upper=\vspace{-6pt},
    before lower=\vspace{-8pt},
    after lower=\vspace{-8pt},
    boxsep=1pt, left=4pt, right=4pt, top=10pt, bottom=10pt,
    center title,
    fonttitle=\scriptsize
  }
}
\tcbset{
  every box/.style={
  }
}
\centering
\begin{minipage}[t]{1.0\linewidth}
  \tcbset{tightsplit,colback=red!5, colframe=red!75!black, title=\ours}
  \vspace{0pt}
  \begin{tcolorbox}
    \scriptsize When rephrasing text in a few words, focus on capturing the main idea and
    key points succinctly. Your summary should be concise, accurate, and free of
    unnecessary details. Aim for clarity and brevity, and ensure that the essence
    of the original text is preserved. Aim for a sentence or two that encapsulates
    the core message.
    \tcblower
    \centering \scriptsize Av ROUGE: $31.60$
  \end{tcolorbox}
\end{minipage}
  
\begin{minipage}[t]{1.0\linewidth}
\vspace{0pt}
  \tcbset{tightsplit,colback=lightyellow!10, colframe=orange!80!black, title=EvoPrompt (GA)}
  \begin{tcolorbox}
    \scriptsize Concisely summarize the main points using clear and
brief language by removing redundancies and
unnecessary details.
    \tcblower
    \centering \scriptsize Av ROUGE: $24.89$
  \end{tcolorbox}
\end{minipage}%

\begin{minipage}[t]{0.49\linewidth}
\vspace{0pt}
  \tcbset{tightsplit,colback=lightyellow!10, colframe=orange!80!black, title=EvoPrompt (GA)}
  \begin{tcolorbox}
    \scriptsize Create a concise summary that retains the semantic
    meaning and omits unnecessary details.
    \tcblower
    \centering \scriptsize Av ROUGE: $29.85$
  \end{tcolorbox}
\end{minipage}
\hfill
\begin{minipage}[t]{0.49\linewidth}
\vspace{0pt}
  \tcbset{tightsplit,colback=cyan!5, colframe=cyan!70!black, title=Manual Instruction}
  \begin{tcolorbox}
    \scriptsize How would you rephrase that in a few words?
    \tcblower
    \centering \scriptsize Av ROUGE: $24.04$
  \end{tcolorbox}
\end{minipage}
\caption{
    Comparison of averaged ROUGE metrics based on prompts generated by \ours{}, EvoPrompt, and Manual Instruction for the summarization task. This figure highlights the importance of precise prompt design: although the two prompts generated by EvoPrompt on two different seeds are superficially similar, they result in significantly different performance. In contrast, the \ours{} prompt is both more effective and better aligned with the task objective.
}
\label{fig:sum_prompt}
\end{figure}

\begin{figure}[ht]
\tcbset{
  tightsplit/.style={
    enhanced,
    before upper=\vspace{-8pt},
    after upper=\vspace{-6pt},
    before lower=\vspace{-8pt},
    after lower=\vspace{-8pt},
    boxsep=1pt, left=4pt, right=4pt, top=10pt, bottom=10pt,
    center title,
    fonttitle=\scriptsize
  }
}
\tcbset{
  every box/.style={
  }
}
\centering
\begin{minipage}[t]{1.0\linewidth}
\tcbset{tightsplit,colback=red!5, colframe=red!75!black, title=\ours}
\vspace{0pt}
\begin{tcolorbox}
\scriptsize Improve the instructions by specifying the aspects of simplification and
providing examples. 
Here's an enhanced version of the prompt: 
Simplify the following text by: \\
1. Shortening sentences. \\
2. Using simpler vocabulary. \\
3. Removing unnecessary or redundant words.\\ 
Example of simplified text: 
Original: The quick brown fox jumped over the fence, which was quite a high
jump for such a small creature. 
Simplified: The quick brown fox jumped over the high fence.
 Your task is to make the given text easier to understand while maintaining its
original meaning.

  \tcblower
 \centering \scriptsize SARI: $57.19$ 
\end{tcolorbox}
\hfill
\end{minipage}

\vspace{-12pt}
\begin{minipage}[t]{0.49\linewidth}
\vspace{0pt}
\tcbset{tightsplit,colback=cyan!5, colframe=cyan!70!black, title=Manual Instruction}
\begin{tcolorbox}
\scriptsize Simplify the text
  \tcblower
 \centering \scriptsize SARI: $43.77$ 
\end{tcolorbox}%
\end{minipage}
\hfill
\begin{minipage}[t]{0.49\linewidth}
\vspace{0pt}
\tcbset{tightsplit,colback=lightyellow!10, colframe=orange!80!black, title=EvoPrompt (GA)}%
\begin{tcolorbox}
\scriptsize Rephrase the given sentence to make it simpler and easier to understand.
  \tcblower
 \centering \scriptsize SARI: $46.60$ 
\end{tcolorbox}
\end{minipage}
\caption{
    Comparison of SARI metric for prompts generated by \ours{}, EvoPrompt and Manual Instruction for the simplification task.}
\label{fig:sim_prompt}
\end{figure}

\newpage

\section{Appendix - Additional Ablation Study Experiments}\label{sec:abl_appendix}
\paragraph{Impact of Prompt Sampling Size in Prompt Selection}

We investigate how the number of prompt samples used during inference in the Prompt Selection technique affects final performance. Specifically, we evaluate \ours{} on the MR classification dataset while varying the number of sampled prompts: $n_{\text{test}} = 1$, 5, 10, and 15. Each configuration is run three times, and we report the average accuracy. The results are presented in Table \ref{tab:abl_test}.

Performance remains stable when using more than five prompt samples, while one prompt only leads to a performance drop. 
%
Although the results suggest that five prompts are sufficient for stable performance, we recommend using ten prompts to provide an additional buffer against potential sensitivity in other tasks or datasets.

\begin{table}[h]
\caption{Model accuracy vs.\ number of test samples}
  \centering
  \small
  {\renewcommand{\arraystretch}{1.2}%
  \resizebox{\columnwidth}{!}{%
    \begin{tabular}{ccccc}
      \toprule
      $n_{\mathrm{test}}$ & 1      & 5      & 10     & 15     \\
      \midrule
      Accuracy     &
        90.92\textsubscript{\textcolor{gray}{±0.17}} &
        91.25\textsubscript{\textcolor{gray}{±0.15}} &
        91.27\textsubscript{\textcolor{gray}{±0.11}} &
        91.35\textsubscript{\textcolor{gray}{±0.11}} \\
      \bottomrule
    \end{tabular}%
  }}%
  \label{tab:accuracy_vs_ntest}
  \label{tab:abl_test}
\end{table}

\paragraph{Influence of Few-Shot Examples}
To ascertain the importance of the automatic inclusion of few-shot examples, we manually remove them from prompt for which PRL provided them and measure performance, see Table~\ref{tab:few-shot-ablation} for results on a subset of classification tasks where PRL produced few-shot examples.
We see that indeed few-shot examples significantly enhance quality.

\begin{table}[h]
\caption{Accuracy on different datasets with and without few-shot learning.}
\centering
\small
{\renewcommand{\arraystretch}{1.2}%
\resizebox{\columnwidth}{!}{%
\begin{tabular}{lcc}
\toprule
\textbf{Dataset} & Acc.\ w/o few-shot & Acc.\ with few-shot \\
\midrule
\textbf{SUBJ} & 66.75 & 77.95 \\
\textbf{CR} & 92.40 & 93.00 \\
\textbf{SST-2} & 95.00 & 96.38 \\
\textbf{MR} & 90.95 & 91.30 \\
\bottomrule
\end{tabular}%
}}%
\label{tab:few-shot-ablation}
\end{table}

\section{Appendix - Sensitivity to the Base Prompt}\label{app:prl_on_better_prompt}
\renewcommand{\arraystretch}{1.5}
In this section, we report results and the corresponding prompts obtained when initializing PRL from an already optimized base prompt, illustrating that performance is largely stable under different prompt initializations.
\begin{table}[h]
\centering
\caption{Effect of base prompt choice on PRL performance (SUBJ).}
\label{tab:base_prompt}
\small
\setlength{\tabcolsep}{6pt}
\renewcommand{\arraystretch}{1.18}
\rowcolors{2}{rowgray}{white}

\begin{tabularx}{\columnwidth}{
  >{\centering\arraybackslash\bfseries}m{1.55cm}
  >{\raggedright\arraybackslash}X
  >{\centering\arraybackslash}m{1.35cm}
}
\toprule
\rowcolor{rowgray}
\textbf{Version} & \textbf{Base prompt} & \textbf{Acc.} \\
\midrule

Manual &
In this task, you are given sentences from reviews. Classify a sentence as \texttt{subjective} if it expresses a personal opinion, or \texttt{objective} if it states a fact. Return only \texttt{subjective} or \texttt{objective}. Consider cues such as \textit{``I think''}, \textit{``it's great''}, \textit{``personally''}, \textit{``impressing''}, \textit{``bad''}, and \textit{``excellent''} as indicative of subjectivity, and treat factual descriptions as objective. &
\textbf{78.55} \\

PRL-based &
In this task, you are given sentences from reviews. Classify a sentence as \texttt{subjective} if it expresses a personal opinion or feeling, and as \texttt{objective} if it states a fact or observation. Return only \texttt{subjective} or \texttt{objective}. Focus on personal viewpoints rather than purely factual content; phrases like \textit{``I think''}, \textit{``it's great''}, and \textit{``personally''} are often subjective cues. &
78.25 \\
\bottomrule
\end{tabularx}
\end{table}

\section{Appendix – Benchmark Results Reproduction}
\label{sec:appendix_reproduction}

To provide a fair comparison with existing benchmarks (APE, APO, EvoPrompt and GRACE), we reproduced their results using Qwen2.5-7B-Instruct as both the Prompt Generator and the Evaluation Model, terms named differently in the original papers but are functionally equivalent. The evaluation procedure is identical for all the benchmarks and for PRL, as follows:

\begin{itemize}
    \item For classification tasks, only a response consisting of a single word denoting the correct class is considered a valid answer.
    \item For summarization and simplification tasks, the entire response generated by the Evaluation Model is used to compute ROUGE/SARI metrics.
    \item For MedQA task, a response including only the proper letter (A, B, C, D or E) is considered a valid answer.
    \item For mathematical reasoning tasks, responses are considered valid if they contain the correct final answer, therefore, models are allowed to produce step-by-step reasoning before concluding.
\end{itemize}

Additionally, we conduct an alternative evaluation protocol for GSM8K in which only the final integer answer is considered valid. Under this stricter setting, PRL continues to achieve the best performance among all methods, as reported in Table~\ref{tab:gsm8k_alternative}.

\begin{table}
\caption{Results on GSM8K with the evaluation protocol accepting only the correct integer.}
    \centering
    \small
    \begin{tabular}{cc}
        \hline
        \textbf{Method} & \textbf{Accuracy} \\ \hline
        MI & 22.20 \\
        APE               & \orangec{27.17}\textsubscript{\textcolor{gray}{±0.65}} \\
        GA    & \yellowc{26.38}\textsubscript{\textcolor{gray}{±1.10}} \\
        DE    & \yellowc{26.38}\textsubscript{\textcolor{gray}{±1.10}}  \\
        PRL       & \redc{29.30}\textsubscript{\textcolor{gray}{±0.05}}  \\
        \hline
    \end{tabular}
    \label{tab:gsm8k_alternative}
\end{table}

To ensure that only the label is output by the Evaluation Model, we appended the following instruction to the end of each prompt in the initial population:

\begin{quote}
\texttt{Return only label \{\textcolor{red}{list of labels}\} without any other text.}
\end{quote}

For example, in the case of the SST-5 dataset, the appended sentence was:

\begin{quote}
\texttt{Return only label 'terrible', 'bad', 'okay', 'good' or 'great' without any other text.}
\end{quote}

In the mathematical reasoning tasks, no additional instruction regarding output is added to the initial prompt in order to enable reasoning. For GSM8K task, the instruction to output only correct integer is added to the initial prompt but only when using the evaluation protocol (1), see paragraph \nameref{par:math_tasks}.

Depending on the benchmark, the initial population consisted of manual prompts (see MI in \nameref{par:benchmarks}) and/or automatic prompts generated by Qwen, following the prompt generation method described in \citep{zhou2022large}.

\paragraph{APO}
Following the EvoPrompt \citep{guo12connecting} experimental protocol, we ran the APO algorithm using the manual prompt with the best performance as the initial population. However, unlike the EvoPrompt authors, we applied the method to all classification tasks, not only binary ones. APO was not evaluated on text generation tasks (i.e., summarization and simplification), as its optimization algorithm fundamentally relies on binary feedback (correct vs incorrect), which is incompatible with continuous scores such as ROUGE or SARI.

The default parameter setup provided by \citep{pryzant2023automatic} was used for each run.

\paragraph{APE and EvoPrompt}
Following \citep{guo12connecting}, the development set size was set to 200 for classification tasks and 100 for simplification and summarization tasks. Each run included 10 iterations. The 10 best prompts for each task served as the initial population (selected from automatic prompts for APE and from both automatic and manual prompts for EvoPrompt).

\paragraph{GRACE} For GRACE \citep{shi2025lossgaingatedrefinement}, we used the default hyperparameters provided by the authors. To reproduce the results on the summarization and simplification tasks, we constructed the binary signal from the top 20\% and bottom 20\% of instances ranked by ROUGE-L/SARI, following the GRACE setup.

\paragraph{RLPrompt}
A direct adaptation of RLPrompt \citep{deng2022rlprompt} to PRL setting was not feasible, since it relies on a fundamentally different evaluation paradigm (selecting the token with the highest probability from a list of predefined verbalizers), much smaller training datasets, and significantly smaller language models. Instead of retraining RLPrompt on Qwen (which would be meaningless in our opinion), we compared the prompts generated by PRL with those reported in the RLPrompt paper, using RLPrompt’s own evaluation method. To ensure fairness despite differences in prompt templates, we evaluated each method with three variants:
\begin{itemize}
    \item sample + prompt: without chat template, 
    \item sample + prompt: with chat template, 
    \item prompt + sample: with chat template
\end{itemize} 
and selected the best score. This approach is justified by the RLPrompt authors’ claim that its prompts exhibit inter-model generalizability. Under this setup, PRL surpasses RLPrompt on every task except SUBJ and achieves a higher average score by 16.5\% (see Table~\ref{tab:rlprompt}).

\begin{table*}[ht]
\caption{Accuracy achieved by prompts from RLPrompt and PRL on classification tasks. \redc{Red} colour mark the best result. The right-most column shows the mean accuracy of each method across the seven datasets.}
\centering
\renewcommand{\arraystretch}{1.2}
\begin{tabular}{cccccccc:c}
\hline
\textbf{Method / Dataset} & \textbf{SST-2} & \textbf{CR} & \textbf{MR} & \textbf{SST-5} & \textbf{AG's News} & \textbf{TREC} & \textbf{Subj} & \textbf{Avg} \\ 
\hline
RLPrompt (2 tokens)
& 65,79 & 58.65 & 72.45 & 40.23 & 70.42 & 38.40 & 68.90 & 59.26 \\
RLPrompt (5 tokens)
& 82.98 & 76.20 & 82.25 & 27.24 & 65.07 & 39.60 & \redc{74.10} & 63.92 \\
\ours{} (ours)
& \redc{95.55}
& \redc{88.60}
& \redc{91.85}
& \redc{56.02}
& \redc{87.39}
& \redc{76.40}
& 67.10
& \redc{80.42} \\
\hline
\end{tabular}%
\label{tab:rlprompt}
\end{table*}

\section{Appendix – Discussions}
\label{sec:appendix_discussions}

\paragraph{Importance of Prompt Optimization}
The recent innovations in instruction tuning, preference alignment, and RLHF reduce prompt brittleness significantly. However, a growing body of the newest work shows that even modern, aligned LLMs still exhibit substantial and often subtle prompt sensitivity, and that explicitly optimizing prompts continues to yield meaningful gains:

\begin{enumerate}
    \item Recent work shows that aligned LLMs remain highly sensitive to small, semantics-preserving prompt changes: 
    \begin{itemize}
        \item \citep{sclar2024quantifyinglanguagemodelssensitivity} find that subtle formatting choices (e.g., bullet style, whitespace, placement of labels) can change few-shot accuracy on LLaMA-2-13B by up to 76 accuracy points, and that this sensitivity persists even with larger models.

        \item \citep{razavi2025benchmarking} systematically vary only the phrasing of prompts with the same information content and show large, unpredictable accuracy swings across state-of-the-art LLMs.
        
        \item \citep{chatterjee2024posixpromptsensitivityindex} and \citep{zhuo2024prosaassessingunderstandingprompt} show that, even for instruction-tuned chat models, the gap between the best and worst semantically equivalent prompts remains large, and that small rephrasings can significantly alter model behaviour.
        
        \item \citep{errica2024did} show that, even when average accuracy is high, minor changes in natural-language instructions often flip predictions.

        \item \citep{cao2024worstpromptperformancelarge} analyze worst-case prompts and show that for models such as Llama-2-70B-chat the difference between best and worst semantically equivalent prompts can exceed 45 percentage points on RobustAlpacaEval, with worst-case accuracy close to random guessing.
    \end{itemize}
    
    \item Small, human-plausible perturbations still hurt robustness:
    \begin{itemize}
        \item \citep{zhu2024promptrobustevaluatingrobustnesslarge} demonstrate that minor typos, synonym substitutions, and light paraphrases, changes that preserve meaning for humans, can substantially degrade performance of LLMs.
        
        \item \citep{li2023evaluatinginstructionfollowingrobustnesslarge} show that short, injected instructions can override original goals in aligned instruction-following models, which implies that specific prompt fragments still exert disproportionate influence despite RLHF.
    \end{itemize}
    
    \item Classical in-context learning results already showed the importance of prompt choice:
    \begin{itemize}
        \item \citep{zhao2021calibrateuseimprovingfewshot} and \citep{lu2022fantasticallyorderedpromptsthem} show that the choice and ordering of few-shot examples can move GPT-style models from near-chance to near–state-of-the-art performance.
        
        \item \citep{min2022rethinkingroledemonstrationsmakes} further show that how demonstrations are written often matters more than their exact labels, reinforcing that prompt design itself is a major performance factor.
    \end{itemize}
    
    \item Even recent large RL-aligned reasoning models exhibit prompt sensitivity:
    \begin{itemize}
        \item The DeepSeek-R1 paper \citep{guo2025deepseek} explicitly notes that a 671B-parameter RL-trained reasoning model remains sensitive to prompt design, and that seemingly reasonable prompting strategies (e.g., few-shot prompting) can decrease performance.
        
        \item The GRACE framework \citep{shi2025lossgaingatedrefinement} further underscores this: it uses DeepSeek-R1 as the optimizer LLM and DeepSeek-V3-0324 \citep{deepseekai2025deepseekv3technicalreport} as the base model for prompt optimization, and shows that purely prompt-level changes yield non-trivial gains over both manual prompts and prior prompt optimizers.
        
        \item GRACE explicitly motivates its design by the “high prompt sensitivity” of the base model and uses DeepSeek-R1’s reasoning ability to craft better prompts for DeepSeek-V3.
    \end{itemize}
    
    \item Our experiments give additional quantitative evidence on aligned models:
    \begin{itemize}
        \item We evaluate PRL on instruction-tuned chat models (Qwen2.5-7B-Instruct, Qwen2-14B-Instruct, Qwen2-32B-Instruct, and LLaMA-3.1-8B-Instruct), i.e., models that already underwent instruction tuning/alignment.
        
        \item Larger evaluation models (Qwen2-14B and Qwen2-32B) still benefit from PRL over strong manual prompts, indicating that prompt optimization remains useful as model capacity grows.
        
        \item Prompts learned on Qwen2.5-7B transfer well to LLaMA-3.1-8B, and retraining PRL directly on LLaMA yields further gains. This suggests that the benefits of prompt optimization are not tied to a single model family.
    \end{itemize}
\end{enumerate}

In summary, the literature shows that even instruction-tuned, RLHF’d, and large LLMs remain highly sensitive to subtle, semantics-preserving prompt variations: this sensitivity affects both average and worst-case performance and is problematic for reliability; and automatic prompt optimization methods do in fact deliver significant gains on modern aligned models.

\paragraph{Path to Universal Prompt Generator}

Retraining the prompt generator for every new task is a clear limitation of the current PRL setup. As we note in the paper, this task-specific regime is standard in automatic prompt engineering, where both prompts and optimizers are typically tuned per benchmark. Nonetheless, moving toward a more universal, task-agnostic prompt generator is an important next step.

Two concrete avenues for the future work are:
\begin{itemize}

    \item Meta-learning/fast adaptation. Instead of fully retraining PRL per task, one could train the generator in a meta-learning regime over many tasks, so that adapting to a new task requires only a small number of RL updates (or even just Prompt Selection). This would retain most of PRL’s performance benefits while greatly reducing per-task cost.
    
    \item Pretraining on large data corpora. Another direction is to run PRL in essentially the same setup as now, but on a large, heterogeneous corpus of tasks and domains, asking the generator to craft prompts for very general settings rather than a single benchmark. The resulting PRL-trained generator would effectively be “pretrained” as a generalist prompter, and could then be lightly fine-tuned with RL on a new target task, improving transfer and reducing the need for full retraining.

\end{itemize}

\section{Appendix – Prompts for Classification}
\label{sec:prompts_cls}
This subsection provides the most effective prompts used for the classification task in our method.

\tcbset{
  prl_prompts_appendix/.style={
    enhanced,
    before upper=\vspace{-8pt},
    after upper=\vspace{-6pt},
    before lower=\vspace{-8pt},
    after lower=\vspace{-8pt},
    boxsep=1pt, left=4pt, right=4pt, top=10pt, bottom=10pt,
    center title,
    fonttitle=\scriptsize, 
    colback=red!5, 
    colframe=red!75!black, 
    title=\ours
  }
}

\begin{figure}[htbp]

\centering
\tcbset{prl_prompts_appendix}
\vspace{0pt}
\begin{tcolorbox}
  \scriptsize In this task, you are to classify the opinion in a given sentence from a review as either subjective or objective.\\
  - A subjective sentence expresses personal feelings, opinions, or attitudes. \\
  - An objective sentence presents facts that can be verified and are not influenced by personal feelings. \\
  Examples: \\
  - Subjective: "This movie is the best I've ever seen." (Opinion expressed) \\
  - Objective: "The movie won five awards this year." (Fact stated) \\
  When classifying, focus only on the opinion, not the facts. Return the label 'subjective' or 'objective' only, without any additional text. Example: \\
  Input: "The food was delicious." Output: subjective
  \tcblower
  \centering \scriptsize Acc.: $77.95$
\end{tcolorbox}

\caption{
    Best prompt generated by PRL for SUBJ classification task along with accuracy.
}
\label{fig:subj_prompt_prl}
\end{figure}

\begin{figure}[htbp]
\centering
\tcbset{prl_prompts_appendix}
\vspace{0pt}
\begin{tcolorbox}
  \scriptsize In this task, you will classify the sentiment of movie review sentences as 'positive' or 'negative'. Examples:\\
    "The movie was thrilling and exciting" -> positive;\\
    "The plot was boring and predictable" -> negative.\\
    Return only the label: 'positive' or 'negative'.
  \tcblower
  \centering \scriptsize Acc.: $96.38$
\end{tcolorbox}

\caption{
    Best prompt generated by PRL for SST2 classification task along with accuracy.
}
\label{fig:sst2_prompt_prl}
\end{figure}

\begin{figure}[htbp]
\centering
\tcbset{prl_prompts_appendix}
\vspace{0pt}
\begin{tcolorbox}
  \scriptsize In this task, you are given sentences from movie reviews. Your goal is to classify each sentence as 'positive' or
'negative' based on its sentiment. Pay close attention to the context and nuances in the text, as the sentiment
might not be explicitly stated. Examples:\\
- "The acting was superb, and the plot was engaging." -> positive\\
- "The movie was so slow and boring that I almost fell asleep." -> negative\\
Return only the label 'positive' or 'negative' without any additional text.
  \tcblower
  \centering \scriptsize Acc.: $93.00$
\end{tcolorbox}

\caption{
    Best prompt generated by PRL for CR classification task along with accuracy.
}
\label{fig:cr_prompt_prl}
\end{figure}

\begin{figure}[htbp]
\centering
\tcbset{prl_prompts_appendix}
\vspace{0pt}
\begin{tcolorbox}
  \scriptsize In this task, you are given a sentence from a movie review. Classify the sentence as 'positive' if the sentiment
is positive, or as 'negative' if the sentiment is negative. Provide only the label 'positive' or 'negative' without
any additional text. Examples:\\
- "The acting was superb and the plot was engaging." -> positive\\
- "The movie was boring and the storyline was predictable." -> negative
  \tcblower
  \centering \scriptsize Acc.: $91.30$
\end{tcolorbox}

\caption{
    Best prompt generated by PRL for MR classification task along with accuracy.
}
\label{fig:mr_prompt_prl}
\end{figure}

\begin{figure}[htbp]
\centering
\tcbset{prl_prompts_appendix}
\vspace{0pt}
\begin{tcolorbox}
  \scriptsize In this task, you are given sentences from movie reviews. Your goal is to classify the sentiment of each
sentence as 'terrible', 'bad', 'okay', 'good', or 'great'.
Be as accurate as possible. Here are the guidelines for each category:\\
- 'terrible': The sentence expresses extreme dissatisfaction or negative feelings.\\
- 'bad': The sentence conveys negative feelings but not as strongly as 'terrible'.\\
- 'okay': The sentence is neutral or has mixed feelings with no strong positive or negative sentiment.\\
- 'good': The sentence conveys positive feelings but not as strongly as 'great'.\\
- 'great': The sentence expresses strong positive feelings or high satisfaction.\\
Consider the overall tone and specific positive or negative words in the sentence to determine the closest
sentiment.\\
If you are not sure, choose the closest option.\\
Return the label 'terrible', 'bad', 'okay', 'good', or 'great' only without any additional text.
  \tcblower
  \centering \scriptsize Acc.: $56.38$
\end{tcolorbox}

\caption{
    Best prompt generated by PRL for SST-5 classification task along with accuracy.
}
\label{fig:sst5_prompt_prl}
\end{figure}

\begin{figure}[htbp]
\centering
\tcbset{prl_prompts_appendix}
\vspace{0pt}
\begin{tcolorbox}
  \scriptsize In this task, you will be given a news article and asked to classify it into one of the four predefined categories: 'World', 'Sports',
'Business', or 'Tech'. \\
Follow these detailed instructions to ensure accurate classification\\
1. Read the article thoroughly to understand its main subject matter.\\
2. Determine which of the following categories the article's main topic most closely aligns with:\\
- 'World': articles covering global news, politics, international affairs, etc.\\
- 'Sports': articles discussing various sports, competitions, athletes, etc.\\
- 'Business': articles focusing on financial news, corporate activities, markets, etc.\\
- 'Tech': articles about technology, innovations, companies, gadgets, etc.\\
3. If the article's content is not clearly related to any of these categories, choose the closest option based on the predominant
subject matter.\\
4. Return the label of the chosen category as a single word without any additional text or explanations, e.g., 'World', 'Sports',
'Business', or 'Tech'.\\
Example: \\Article: "Apple Launches New iPhone Model with Improved Camera Features" Label: Tech\\
Article: "China and the US Reach a New Trade Agreement" Label: World\\
Article: "Local Soccer Team Qualifies for the World Cup" Label: Sports\\
Article: "Oil Prices Drop as OPEC Decides to Cut Production" Label: Business
  \tcblower
  \centering \scriptsize Acc.: $84.42$
\end{tcolorbox}

\caption{
    Best prompt generated by PRL for AG's News classification task along with accuracy.
}
\label{fig:news_prompt_prl}
\end{figure}

\begin{figure}[htbp]
\centering
\tcbset{prl_prompts_appendix}
\vspace{0pt}
\begin{tcolorbox}
  \scriptsize Please perform a Question Classification task. Given a question, classify it into one of the following categories:\\
- **Description**: Questions asking for descriptions or explanations.\\
- **Entity**: Questions asking about specific things, objects, or entities.\\
- **Expression**: Questions asking about how something is expressed or phrased.\\
- **Human**: Questions asking about people, their characteristics, or roles.\\
- **Location**: Questions asking about places or geographical locations.\\
- **Number**: Questions asking for numerical information or quantities.\\
Return the label 'Description', 'Entity', 'Expression', 'Human', 'Location', or 'Number' only without any additional text.
Example: \\Question: "What is the capital of France?" Label: Location\\
Question: "How do you say 'hello' in Spanish?" Label: Expression\\ Question: "Who is the CEO of Apple?" Label: Human
  \tcblower
  \centering \scriptsize Acc.: $78.60$
\end{tcolorbox}

\caption{
    Best prompt generated by PRL for TREC classification task along with accuracy.
}
\label{fig:trec_prompt_prl}
\end{figure}

\section{Appendix - Usage of LLMs}
LLMs were used for editorial support, including polishing the manuscript’s writing and presentation, and for help drafting some implementation code. The authors made all substantive research decisions and contributed all core technical ideas.

\end{document}